\documentclass[IJPHM, 2022, 00]{PHMSociety}
\usepackage[utf8]{inputenc}
\usepackage{comment}
\usepackage[english]{babel}  
\usepackage{natbib}
\usepackage{graphicx}
\usepackage{soul}
\usepackage{pythontex}
\usepackage{url}
\usepackage{array}
\usepackage{tabularx}
\usepackage{xcolor}
\newcommand\todo[1]{\textcolor{blue}{TODO: #1}}
\usepackage{makecell}
\usepackage{caption}
\usepackage{subcaption}

\begin{document}

\title{Technical Language Supervision for \\ Intelligent Fault Diagnosis in Process Industry}

\author{%
	Karl Löwenmark\authorNumber{1}, Cees Taal\authorNumber{2}, Stephan Schnabel\authorNumber{3}, Marcus Liwicki\authorNumber{4}
	 and Fredrik Sandin\authorNumber{5}
}

\address{
	\affiliation{{1,4,5}}{Embedded Intelligent Systems Laboratory (EISLAB), Luleå University of Technology, \\971 87 Luleå, Sweden}{ 
		{\email{karl.ekstrom@ltu.se}}
        {\email{marcus.liwicki@ltu.se}}
        {\email{fredrik.sandin@ltu.se}}
		} 
	\tabularnewline 
	\affiliation{2}{SKF Research \& Technology Development, Meidoornkade 14, 3992 AE Houten, \\P.O. Box 2350, 3430 DT Nieuwegein, The Netherlands}{ 
        {\email{cees.taal@skf.com}}
		} 
	\tabularnewline 
	\affiliation{3}{SKF Condition Monitoring Center Luleå AB, 977 75 Luleå, Sweden}{ 
        {\email{stephan.schnabel@haw-landshut.de}}
		} 
}

\maketitle
\pagestyle{fancy}
\thispagestyle{plain}

\phmLicenseFootnote{Karl Löwenmark}

\begin{abstract}

%
%
In the process industry, condition monitoring systems with automated fault diagnosis methods assist human experts and 
thereby improve maintenance efficiency, process sustainability, and workplace safety.
%
Improving the automated fault diagnosis methods using data and machine learning-based models is a central aspect of 
intelligent fault diagnosis (IFD).
A major challenge in IFD is to develop realistic datasets with accurate labels needed to train and validate models, and to transfer models trained with labeled lab data to heterogeneous process industry environments.
However, fault descriptions and work-orders written by domain experts are increasingly digitised in modern condition monitoring systems, for example in the context of rotating equipment monitoring.
%
Thus, domain-specific knowledge about fault characteristics and severities exists as technical language annotations in industrial datasets.
Furthermore, recent advances in natural language processing enable weakly supervised model optimisation using natural language annotations, most notably in the form of \textit{natural language supervision} (NLS).
This creates a timely opportunity to develop \textit{technical language supervision} (TLS) solutions for IFD systems grounded in 
industrial data, 
for example as a complement to pre-training with lab data to address problems like overfitting and 
inaccurate out-of-sample generalisation.
%
%
%
%
%
%
%
%
We surveyed the literature and
identify a considerable improvement in the maturity of NLS over the last two years, facilitating 
applications beyond natural language; a rapid development of weak supervision methods;
and transfer learning as a current trend in IFD which can benefit from these developments.
%
%
%
Finally we describe a general framework for TLS and implement a TLS case study based on SentenceBERT and contrastive learning based zero-shot inference on annotated industry data.
\end{abstract}

\section{Introduction} 
\begin{table*}[t]
\centering
\caption{Fault Diagnosis Tasks}
\begin{tabular}{||l|| c | c | c| c||} 
  \hline
Task & Question addressed  & Output & Added value  & Automated? \\
 \hline 
 \hline
Detection & Is there a fault present? & Yes/No & Alert human analysis & Yes\\ 
 \hline
Classification & What type of fault? & Class & Guide human analysis & Partially\\
 \hline
Severity & How severe is the fault? & Magnitude & Motivate maintenance & No\\
 \hline
RUL & Time until maintenance is needed? & Risk vs Time & Maintenance planning & No\\
 \hline
Root Cause & What caused the fault? & Description 
& Preventive policies 
& No
\\
 \hline
\end{tabular}
\label{tab:tasks}
\end{table*}
Condition-monitoring (CM) based fault diagnosis of rotating machinery \citep{condition_monitoring, machinery_diagnostics_review} is widely used in industry to optimise equipment availability, uniformity of product characteristics and safety in the work environment, and to minimise production losses and material waste.
In process industry, this typically requires human expert analysts with years of training and detailed knowledge about the operational states, functional roles and contexts of the machines being monitored.
Due to growing demands on production efficiency and the vast amounts of data consequently generated in modern CM systems, automated fault diagnosis systems \citep{Kothamasu2006} are required to assist human analysis through alarms and policy recommendation.
Important tasks for the automated system are fault detection and classification to generate alarms and filter data, and fault severity estimation to predict remaining useful life and recommend policy options.
Existing automated systems are mainly based on expert systems \citep{knowledge_based_recommender_system}, with a knowledge-base derived from physical properties of analysed components, and a rule-based inference engine with local thresholds set by experts \citep{SKF_Observer}.
In the case of vibration measurements of rotating machinery, signal processing and kinematics based condition indicators are commonly used as knowledge-bases \citep{signal_processing_2006,signal_processing_2016, diagnostics}.
Intelligent fault diagnosis (IFD) \citep{roadmap} has been proposed to enhance the automated systems by inferring fault characteristics directly from process or lab data through learning based methods.
Improving existing models is vital to meet the increasing demands on CM systems to improve production and equipment life cycle efficiency in process industry \citep{processit,CBM_review}, and the machine CM market is estimated at \$2.6 billion with a compound annual growth rate estimation of 7.1\%\footnote{\url{https://www.marketsandmarkets.com/Market-Reports/machine-health-monitoring-market-29627363.html}}.
For example, improved IFD algorithms can contribute to: reducing the number of unnecessary interventions; facilitating remanufacturing of components \citep{skf_report_2020}; optimising maintenance schedules; and enabling analysts to focus on qualified preventive tasks.

However, it is difficult to develop realistic datasets with accurate labels needed to train and validate IFD models, and such data are expected to generalise poorly between process-industry plants due to their heterogeneous nature.
Recent innovations in natural language processing offer a timely opportunity to address this challenge with methods used in \textit{natural language supervision} (NLS)  \citep{NLS_review} using digitalised technical language fault descriptions and work-orders available in many process industry datasets.
Processing technical language poses unique challenges different from natual language, promoting the need for \textit{Techincal Language Processing} (TLP) and a technical version of NLS in \textit{Techincal Language Supervision} (TLS).
Therefore we survey the state of the art in IFD, NLS and TLP, and discuss how TLS can be applied to IFD in a process industry context.

\subsection{Background}
Fault Diagnosis (FD) deals with the mapping of measured signal features to component conditions. 
The most basic condition is whether a fault is present or not, but more complex estimations such as fault class, fault severity, remaining useful life (RUL) and root cause analysis (RCA) can also be required. 
Table \ref{tab:tasks} describes these five major subtasks of FD, ordered in rising complexity based on interviews with condition monitoring experts from process industry. 
Fault detection and classification are tasks that are frequently automated in process industry through signal processing \citep{Kothamasu2006, knowledge_based_recommender_system, SKF_Observer}, and for example model-based thresholding.
Fault severity estimation, a vital tool in maintenance decisions, is next in line to be automated, but is challenging due to nonlinear relationships between signal features and fault evolution \citep{Rolling_bearing_FSE}.
RUL depends on the evolution of estimated fault severity over time, and predicts the remaining time until a fault is so severe that a component is no longer useful \citep{RUL_health_review, RUL_review}.
RCA is a complex task that may be challenging to automate, but will indirectly be improved if simpler tasks are automated and human experts can invest more time in preventive policies.

\begin{figure*}[h!]
    \centering
    \includegraphics[width=0.9\textwidth]{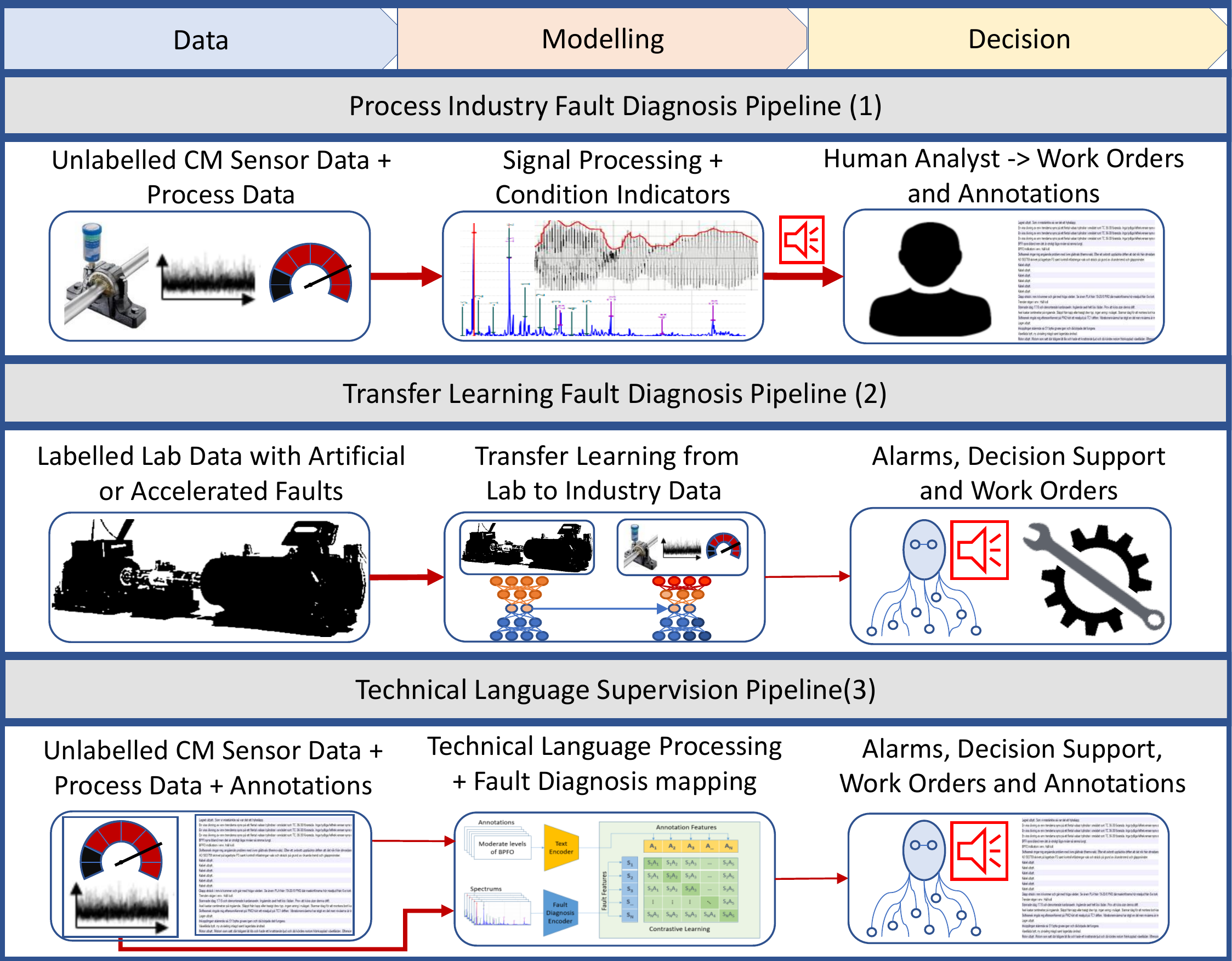}
    \caption{An overview of a typical process industry fault diagnosis pipeline (1), possible transfer learning IFD pipeline additions (2), and our suggested natural language supervision pipeline (3). Both (2) and (3) can provide considerable contributions to (1), with the strongest contributions coming from both pipelines implemented in symbiosis.
    }
    \label{fig:new_pipeline_CM}
\end{figure*}
%
%
The upper part of Figure~\ref{fig:new_pipeline_CM} illustrates an example of a typical FD system (labeled Pipeline~1) implemented in process industry, see for instance \citep{SKF_Observer, pdm_VA, emerson_kinematics}.
The system requires no fault history data to learn from, but requires process information and kinematic models for the extraction of condition indicators \citep{condition_indicators_2016}.
Faults are detected and classified using signal processing \citep{machinery_diagnostics_review}, for instance root mean square, peak-to-peak and time synchronous average in the time domain \citep{time_series_data_mining}; spectral density, enveloping and Hilbert transform in the frequency domain; and dictionaries, wavelets and the Wigner-Ville distribution in the time-frequency domain; as well as kinematics based condition indicators, for instance the frequency intensity in the ball pass frequency of the outer race in ballpoint bearings. The decomposed signal is then analysed with typically simple rules based on indicator magnitude defined by experienced analysts.
Once a fault is detected by the model, a human analyst is alerted for in-depth diagnosis.
The analyst decides whether to further investigate alarms or not, describes eventual faults in the form of natural-language annotations and makes work orders.
Thus, the automated FD model acts like a filter between the massive amount of sensor data that is constantly generated, and the accurate but resource-constrained analysis of human experts.
Based on cases from two industry collaborations with major process industry actors in Northern Sweden\footnote{Smurfit Kappa, 700 000 tonnes of Kraftliner per year, and SCA Munksund, 400 000 tonnes of Kraftliner per year}, analysts monitor around 5000 alarms per analyst per year, after filtering, where at most 20\% of generated alarms point to component faults and the rest are due to temporary or constant signal malfunctions.

With improved automated FD, analysts could focus on more advanced fault diagnosis tasks beyond the current capabilities of IFD.
Considerable research has been invested in automated FD, and many learning-based methods have shown promising results on test datasets \citep{AIFD_review_2018, ML_WTCM_2019, survey_DLFD_2019}.
However, the accurate deep learning models used in many IFD publications require vast amounts of training data in the form of labelled datasets, sets that typically do not exist in process industry cases \citep{DL_SHM_2018}.
Instead, training and test datasets are created in lab environments with artificial or accelerated fault development, such as the Case Western Reserve University bearing dataset \citep{cwru_webpage}, the Intelligent Maintenance System (IMS) by the University of Cincinnati dataset \citep{IMS_nasa_webpage}, and the Machinery Failure Prevention Technology (MFPT) dataset \citep{MFPT_webpage}, but typically generalise poorly to heterogeneous environments \citep{CWRU_review, zhang2019semisupervised} such as process industries.
%
Thus, despite the maturity of IFD methods in terms of literature, supervised IFD lacks wide-spread implementation in industry.

Industry datasets suitable for IFD can in some cases potentially be created, but it is difficult and costly to define high-quality labels that are accurately connected to relevant data.
Therefore, transfer learning \citep{SCHWENDEMANN2021104415}, illustrated in Pipeline~2 in Figure~\ref{fig:new_pipeline_CM}, has become an increasingly popular approach to develop IFD methods without requiring a large labelled dataset in the target domain \citep{roadmap}.
Ideally, a model could be developed/trained with data from a lab environment, then transferred to similar components in an industrial environment.
However, this remains a challenging goal due to differences between developing faults, heterogeneous environments, varying sensor and signal-to-noise conditions, and complex coupling of signal components.
Thus, a method for the extraction of labels for industry data would be valuable and can facilitate implementations of current IFD models, as well as transfer learning by providing access to labels in the target domain.

While labels are lacking in realistic CM datasets, technical language fault descriptions are written by analysts when documenting and monitoring the development of for example bearing faults over long periods of time (several months).
Thus, the text-annotations produced as outputs of Pipeline~1 in Figure~\ref{fig:new_pipeline_CM} contain valuable albeit noisy information about fault development characteristics and severities.
This motivates the question, can such domain-specific annotations and related knowledge be used for training and fine-tuning of IFD methods as a substitute for regular labels?

Language has been used to train machine learning models for image recognition and object detection through recent breakthroughs in natural language supervision \citep{radford2021learning, ramesh2021zeroshot}.
Can a similar approach be used to train IFD models on industry data using annotations and work orders as zero-shot labels?
\subsection{Contribution}

We propose the usage of TLS on technical language fault descriptions to overcome the lack of labels in industry CM datasets.
TLS is grounded in three fields, IFD, TLP and NLS, and we briefly survey all three to motivate the purpose and benefits of TLS.
Potential TLS contributions supervision are improved support for human analysts and automation of simpler tasks by augmenting the label domain for transfer learning or zero-shot learning.

Pipeline~3 in Figure~\ref{fig:new_pipeline_CM} illustrates the concept of a technical language supervision framework for process industry data.
Unlabelled CM sensor data and process data are used to extract features through methods already used in IFD models, and the features are mapped to annotation embeddings.
In the implementation stage, an unannotated signal is thereby mapped to the closest language fault queires in the joint embedding space, and with a sufficiently good model and well chosen queries, the fault class and severity can be estimated and described.
Besides alarms and work orders, a language based model could also retrieve spectra from queries and generate new annotations and descriptions of detected faults.
We implement a TLS model based on process industry signals and annotations, and show an example of spectrum retrieval from free form queries, as well as zero-shot fault classification of spectra.
%
%
%

\begin{figure}[h]
\centering
\includegraphics[width=0.5\textwidth,trim=0 0 0 8mm, clip]{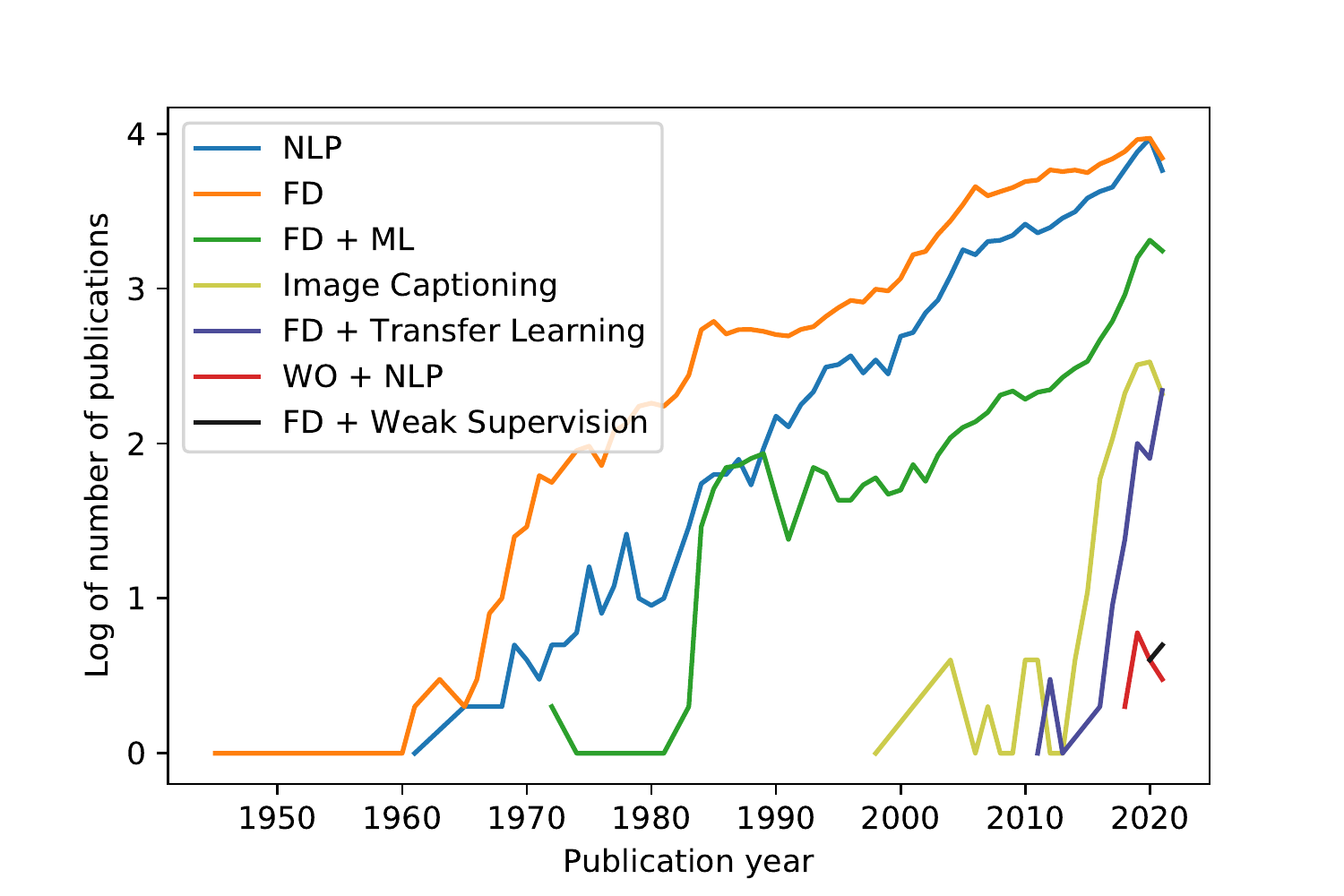}
\caption{
Trends of publications between 1967 and 2020, obtained through Scopus queries looking for publications with the targeted keywords in the article title, the abstract or the keywords.
For instance, a query for fault diagnosis related keywords and transfer learning is designed as follows:
\textit{("condition monitoring"  OR  "fault diagnosis"  OR  "fault classification"  OR  "fault detection" )  AND  "transfer learning"}
The annual number of articles about the application of machine learning (ML) to condition monitoring (CM) and fault diagnosis (FD) increases exponentially.
That is also the case for the annual number of natural language processing (NLP) articles, which now equates the total annual number of FD-related articles.
A total of 15 articles that use NLP on work orders (WO) were found, but no implementations of natural language supervision on fault diagnosis problems were identified.
Weak supervision, or weakly supervised learning, is also not yet commonly used, with 4 articles in 2020 and 5 articles so far in 2021. 
%
}
\label{fig:1}
\end{figure}

\begin{table*}[t]
\caption{Fault Diagnosis Model Frameworks}
\centering
\begin{tabular}{||l|| c | c | c||} 
  \hline
Framework & Data requirements & Challenges \\
 \hline 
 \hline
Unsupervised Learning & CM data & Applications beyond fault detection \\ 
 \hline
Supervised Learning & Labelled CM dataset & Lack of labelled industry data \\ 
 \hline
Transfer Learning & Labelled lab dataset, CM data & Lab features different from industry features \\ 
 \hline
Weak Supervision & Weakly Labelled CM dataset & Still requires labels \\ 
 \hline
Language Based & CM dataset, CM annotations & Not yet applied in IFD \\
 \hline
\end{tabular}
\label{tab:models}
\end{table*}

\subsection{Research Trends}
We also surveyed the fault diagnosis literature and recent publications on language-based learning in the context of natural language supervision and image captioning to identify the trends of publications that combine these concepts.
Figure~\ref{fig:1} shows the number of published articles per year according to Scopus for search queries containing keywords related to fault diagnosis and machine learning. 
%
We present the publication trends of natural language processing (NLP), fault diagnosis (FD), fault diagnosis with machine learning (FD + ML), fault diagnosis with transfer learning (FD + Transfer Learning), image captioning, work orders with natural language processing (WO + NLP), and finally fault diagnosis with weak supervision.
For fault diagnosis, a query including "condition monitoring"  OR  "fault diagnosis"  OR  "fault detection" OR "fault classification"  was used.
For machine learning, "machine learning"  OR  "data driven"  OR  "deep learning"  OR  "artificial intelligence" were used.
The queries "transfer learning", "work order", "natural language processing" and "image captioning" were used explicitly as is.
Weak supervision was queried as "weak supervision" OR "weakly supervised".

The trends show that machine learning is increasingly applied in the FD literature, and that transfer learning has become increasingly popular, going from 2 publications in 2016 to 178 publications in 2020.
NLP is a rapidly evolving field of research, with significant practical advancements in the last decade.
This is also reflected in the swift growth of image captioning publications starting in 2015, increasing from 11 to 322 publications in four years.
%
15 publications that use natural language processing with work orders were found, but NLP was employed for information retrieval, and no publications combining natural language supervision with IFD were found.
%
%
Weak supervision only appeared nine times (with one valid article scheduled for 2022 not counted) in our queries, but notably three articles were cited more than ten times; \cite{weak_supervision_IFD_1} with 64 and 30 \citep{weakly_supervised_clustering} citations, and \cite{weakly_supervised_2021} with 12 , showing that the interest far outweighs the current publication number.
Articles citing weak supervision articles were mainly focused on transfer learning, but we predict an increase in direct mentions of weak supervision methods.


\subsection{Outline of article}

In Section II, we describe the application of FD in process industry, which is subject to constraints related to the high cost of unplanned stops that can affect the whole production process.
Five principal FD tasks are described, and the related methods and algorithms used for automated FD are also presented.
In section III we briefly review natural language supervision and related fields such as image captioning, and discuss how natural language can be integrated in an IFD framework.
Section IV describes a case study implementation of a TLS solution for IFD based on theories from section III, using process industry data for training and illustrations of model performance.
We focus on rotating machinery in process industry, but in principle the framework of technical language supervision is expected to generalise to other fault diagnosis applications where fault descriptions are also present.

\section{Deep Learning in Intelligent Fault Diagnosis}

Table \ref{tab:models} summarises different 
data-driven methods used for IFD, besides the kinematic rule-based method already discussed in the background.
The methods are ordered roughly by maturity and data requirements.
Unsupervised learning applies directly to unlabelled CM data, and it is partially implemented in process industries \citep{SKF_Enlight_AI, amazon_monitron, amazon_monitron2, emerson_ML}.
Supervised learning requires a labelled dataset in the application environment, and is widely investigated in the literature \citep{basic_data_driven_IPM, DL_SHM_2018, AIFD_review_2018, DL_wind_turbines, ML_WTCM_2019, DL_bearing_faults}, but not in process industry.
Transfer learning requires a labelled dataset for pre-training, and data from the application environment, ideally labelled, for fine-tuning.
The number of articles on transfer learning has increased rapidly in the last decade, but although transfer between lab environments show great results, we find no articles that apply transfer learning methods directly on process industry data.
Finally, natural language supervision based learning only requires unlabelled CM data with associated annotations, but this method remains to be adapted and investigated for fault diagnosis tasks.
The first mentions of natural language processing for in an IFD context was 2020 in the name of "technical language processing", though natural language supervision is yet to be introduced to IFD.

\subsection{Unsupervised Learning}

Unsupervised learning, i.e learning patterns without labels, is connected to the modelling module of Figure \ref{fig:new_pipeline_CM} and is primarily used for \textit{clustering}, \textit{encoding}, \textit{feature extraction} and \textit{anomaly detection} \textit{fault detection} \citep{fault_detection_unsupervised}.
%
Models commonly used for clustering are k-means, Principal Component Analysis (PCA) and t-distributed Stochastic Neighbor Embedding (t-SNE).
Auto-Encoders and variational auto-encoders \citep{fault_detection_AE, IFD_VAE_extreme} and Dictionary Learning \citep{dictionaries_elad, dictionaries} are commonly used for Encodings and Anomaly Detection.
Virtually all models can be used to reduce dimensionality and extract features depending on the data, with PCA and t-SNE being more direct dimensionality reductions and auto-encoders serving as a more complex reconstruction model, often with encoders/decoders based on convolutions, recurrence or transformers.

Clustering, encodings and feature extraction can be valuable ways of understanding, simplifying or visualising data.
A CM dataset with healthy and unhealthy data can with the right methods and data be divisible in to two clusters, which can then be manually labelled healthy and unhealthy, thus detecting faults \citep{k_means_clustering_detection}.
Likewise, encodings or extracted features can serve as values in a simple rule-based system for fault detection or classification, and extracted features in particular can give valuable insight in feature importance.
Regardless, the lack of a supervision signal necessitates a human in the last step to validate or assign meaning to clusters, encodings or features, before the model is ready to automatically detect faults.
Anomaly detection can work more autonomously by learning the healthy state of a signal, then classifying deviations from this state as detected faults \citep{MARTINDELCAMPO2017187} or into fault classes \citep{FD_rotating_machinery_AE}.

However, healthy states that lack sufficient presence in the training set has a risk of being classified as unhealthy at deployment, and unhealthy states that are present during training might be considered healthy, which is difficult to detect due to the lack of labels in the dataset.
Furthermore, deviations might occur due to healthy states, or in directions relatively orthogonal to previous deviations.
Such issues fall within the scope of \textit{zero-shot learning}, wherein a model is required to observe and predict samples from a previously unseen class or distribution.\citep{zero_shot_review_2021}.
For zero-shot learning to work, there has to be a distinct characteristic of faults and healthy states that is true for previously unseen faults or healthy states, which can be leveraged to assign these distributions to the correct class.
NLS is sometimes discussed in the scope of zero-shot learning, and zero-shot learning techniques are often used in NLS.
Likewise, zero-shot learning can be used to augment supervised learning methods beyond classes present in the supervision signal, but it is best described under the umbrella term of unsupervised learning or through the lens of weak supervision, as discussed in Section \ref{sec:weak_supervision}.

\subsection{Supervised Learning}

Supervised Learning can be employed for any FD task, as long as sufficient data and good labels are present.
Transfer Learning, Weak Supervision and Language Supervision are all arguably subgroups of supervised learning explicitly designed to circumvent the limitation of requiring good labels.
Architectures used in supervised learning are thus also employed in its derivatives, though with different learning procedures, just as how for instance auto-encoders from unsupervised learning can be used together with an output layer in a supervised paradigm.

Supervised learning architectures used in IFD range from shallow models such as tree-based models, e.g. random forest \citep{Random_forest_1}; support vector machines \citep{basic_data_driven_IPM, data_driven_ipm}; probabilistic models such as Bayesian statistics \citep{bayes_1, bayes_NN}; and deep architectures such as fully connected feed forward deep neural networks \citep{fault_detection_DNN}; (variational) auto-encoders with classification layers \citep{VAE_IFD, IFD_VAE_extreme}; convolutional neural networks \citep{fault_classification_CNN, fault_classification_liftingnetcnn}, commonly used in image analysis; recurrent neural networks \citep{RNN_2018, RNN_2020, RNN_2021}, commonly used in language analysis but applicable on sequential data in general. Importantly, supervised learning has been employed for fault severity estimation \citep{Rolling_bearing_FSE} and RUL prediction \citep{RUL_CNN,RUL_CNN_2,RUL_ANN,RUL_RNN,RUL_health_review,RUL_review}.

Labelling industry datasets for supervised learning can facilitate implementations in that industry environment, but the labelling process is costly, and requires analyst efforts.
Furthermore, some faults have stochastic features, for example due to the varying nature of the source geometry or signal transfer function, and are thus difficult to generalise with supervised classifiers.
In general, faults are undesirable and therefore relatively scarce in industrial datasets, but are required in training datasets for supervised learning.
Consequently, producing a labelled industry dataset for supervised learning would require considerable resources and potentially occupy analyst time necessary for condition monitoring. 
Therefore, fault classification models described in the literature are typically trained on labelled data from lab environments, where faults are generally either artificially induced or provoked through intense loads, as it might take several years until faults develop naturally.
The development of the fault is then accelerated by e.g high loads or starved lubrication, which increase fault development per revolution, and high speeds to increase revolutions per minute (RPM).
High RPM also produce higher signal-to-noise ratios as some noise is stationary and fault features increase more in magnitude than noise features.

Ideally, a model supervised on a component in a lab environment would then be deployable in an industry environment, but there are two issues that makes this difficult.
Firstly, artificial or accelerated fault developments result in fault characteristics that are different compared to faults in industry environments.
Therefore, the decision boundaries do not necessarily generalise well from lab to industry environments, and the feature space can differ due to different fault development processes.
Secondly, signals generated in a lab setting differ greatly from signals in an industry environment where a component is connected to several other components in a larger system, and signal components are combined and masked by noise.
The signal to noise-ratio will consequently be lower in the industry environment, and the coupling with surrounding components can shift the true feature space as well.
Thus, direct supervised learning works best in the environment where it has been trained, and generalisation can be difficult unless labels are preserved in the target space.

\subsection{Transfer Learning}
\label{sec:transfer_learning}
Recently, the research focus in IFD has shifted to include methods to overcome the lack of labels in industry datasets such as transer learning and weak supervision.

Transfer learning seeks to develop methods for training of a model in one environment, then fine-tuning the feature space and decision boundaries to suit implementation in another environment \citep{transfer_learning_review}.
In situations with sparse data optimisation limits, transfer learning can use domains with rich data, such as lab datasets, to infer necessary knowledge \citep{Zhang2021_recommender}.
The research on transfer learning in fault diagnosis applications has increased rapidly over the last few years, with many successful transfers between different lab datasets \citep{roadmap}.
As models improve, transfer learning can enable broader implementation of these models in process industry with a lower demand for labeled instances compared to supervised learning \citep{transfer_learning_small_dataset_CNN_2018}, while solving the same tasks.

%
Methods used in transfer learning vary;
many publications use transferrable convolutional neural networks \citep{transfer_learning_preprocessingfree_CNN_2018, transfer_learning_deep_2019, transfer_learning_lab_to_locomotive_2019, transfer_learning_deep_convolutional_2019, transfer_learning_CNN_gas_turbine_2019, transfer_learning_online_CNN_2020, deep_transfer_learning_2020, transfer_learning_ensamble_CNN_2020, transfer_learning_resnet_2020, transfer_learning_rotary_machinery_CNN_2020, transfer_learning_varying_workcond_CNN_2021}, occasionally employed with adversarial networks \citep{transfer_learning_GAN_2019, GAN_IFD, transfer_learning_GAN_2020}; some use recurrent neural networks \citep{transfer_learning_recurrent_2018, transfer_learning_RNN_arbitrary_length_2019, transfer_learning_RNN_2020}; auto-encoders are also used \citep{transfer_learning_sparse_auto_encoder_2019}, and recently weak supervision \citep{weakly_supervised_transfer_learning} and digital twin-based transfer learning \citep{digital_twin_IFD} have been successfully implemented.

Transferring knowledge from one environment to another adds an additional benefit to symbol-feature relation graphs besides illustrating the process of the reasoning module.
Humans learn concepts in a highly transferable manner, and it is for instance highly feasible that an experienced analyst could diagnose faults in a previously unseen environment with good accuracy, while a learning based model would certainly fail at adapting unless optimised through transfer learning.
The underlying concepts of fault developments are likely the same in both environments, which is what humans use to generalise knowledge.
Optimizing not only direct mappings, but symbol-feature relation graphs as well, can thus create models with stronger generalisability by mimicking human knowledge \citep{tenenbaum1}.

\subsection{Weak Supervision}
\label{sec:weak_supervision}

\begin{table*}[h!]
    \centering
    \caption{Different weak supervision challenges, causes and solutions}
    \begin{tabular}{||l|| c | c ||} 
        \hline
        Weak supervision group & Cause & Solution \\
         \hline 
         \hline
        Incomplete labels & Missing labels from datapoints & \makecell{Active learning \\ Semi-supervised learning \\ Few-shot learning \\ Zero-shot learning} \\ 
         \hline
         Inexact labels & Multiple datapoints per label &\makecell{Multi-instance learning \\ Contrastive learning} \\ 
         \hline
        Inaccurate labels & Label is wrong & \makecell{Regularisation \\ Re-labelling} \\
         \hline
        \hline
    \end{tabular}
    
    \label{tab:weak_supervision}
\end{table*}

Weak supervision is an umbrella term for a set of methods developed to perform supervised tasks on data where labels are insufficient for regular supervised learning \citep{weaklySupervised}.
It can work in conjunction with transfer learning to enhance fine-tuning on the target dataset, or stand-alone to facilitate direct optimisation in the target environment.
Table \ref{tab:weak_supervision} illustrates three major ways in which labels can be insufficient, the cause, and proposed methods to amend the issue.

\subsubsection{Incomplete supervision}
\textit{Incomplete labels} are characterised by a dataset where most data points are unlabelled.
In a CM dataset, faults that have not been discovered yet are a cause for incompleteness, as this prevents the assumption that all unlabelled data is healthy data.

The main strategy for dealing with incomplete datasets is called \textit{semi-supervised learning} \citep{semi-supervised-survey, self-semi-supervised, self_supervised_EAoAI}, which aims to create clusters of features that correspond to the available labels, and to estimate the probability that an unseen feature belongs to one of the identified clusters.
Semi-supervised learning has been employed in IFD settings on lab datasets with partial \citep{semi_supervised_surface_estimation} or limited labels \citep{semi_supervised_IFD}.
By implementing semi-supervised learning on a CM dataset with natural language supervision, it is possible to include all time series data for a prediction, where particularly noisy samples would be less likely to affect the model optimisation process, as they are likely distributed far away from the cluster centres.
The diagnosis of faults in unlabelled samples also belong to the domain of semi-supervised learning, albeit with the additional challenge associated with many unique components and features.
This challenge can necessitate \textit{active learning} \citep{aghdam2019active, self_supervised_EAoAI}, in which a model identifies selected unlabelled datapoints and alerts a human expert to label them.
Active learning requires human intervention, but aims to make use of human efforts as efficiently as possible to improve the model accuracy \citep{Zhang2021_recommender}.
Another scheme used to overcome incomplete labels is few-shot learning \citep{few_shot_survey_2020}, where a model is optimised to perform supervision tasks with insufficient data for normal supervision training \citep{few_shot_SPARC_2018, few_shot_bearings_2019, few_shot_IFD_2020}.
Few-shot learning provides an interesting opportunity to learn fault features with only a few instances in a training datasets, as can be the case for many rare faults or components.
In the case where no labels exist, supervision algorithms might still be applicable through zero-shot learning \citep{zero_shot_review_2021}.
In zero-shot learning, the model seeks to generalise knowledge from seen classes to unseen classes with similar behaviour, much like how humans can see images of house-cats and dogs and then correctly categorise lions to felines and wolves to canines \citep{zero_shot_2020, zero_shot_2021}.

\subsubsection{Inexact supervision}
\label{sec:inexact}
\textit{Inexact labels} coarsely describe some aspects of the ground truth for a set of features, but do not accurately define it.
In general, symbols like labels can not fully represent physical processes of unknown dimensions.
Instead, labels define semantics at a certain level of approximation and scale.
Thus, labels of physical processes are by nature incomplete semantical descriptions of reality.
CM annotations do not describe the properties of each recording in a faulty component, only that from a large bag of recording a fault has been diagnosed.
The fault features from each recording were likely not equally important for the diagnosis however, and learning which features imply which diagnoses would be easier if each recording had its own label.
%

The challenges of inexact labels have been proposed to be overcome through multi-instance learning \citep{multi_instance_one, hoffmann-etal-2011-knowledge, zeng-etal-2015-distant} and contrastive learning.
In multi-instance learning, the optimisation algorithms seeks to find the common denominators in the label "bags" that are present for learning.
By learning from which components were replaced and which were not, correlations in underlying features such as fault severity or deterioration speed can be associated as parts of the bag and used for predictions.

%
%
%
%
%
%

\subsubsection{Inaccurate supervision}
\textit{Inaccurate labels} occur when analysts make fault diagnosis mistakes.
This is unlikely to occur with fault classification, but possible with fault severity due to the higher complexity of that task.
An analyst can for example assume that a fault may be severe and order a replacement of the component to avoid failure, while the fault actually is minor.

Inaccurate labels are characterised by not conforming to the ground truth, in other words being wrong.
To learn with noisy or inaccurate labels, a model seeks to identify and potentially correct incorrect labels \citep{noisy_labels_2018}.
Thus, the model maintains some trust in its predictions, capable of deeming the label inaccurate when confidence in prediction is high and label features deviate from similar labels \citep{noisy_labels_2019}.
This trust can be reinforced with physics induced machine learning to maintain a baseline estimate of how labels and signals should correlate, based on physical knowledge of the problem.

%

\section{Technical Language Supervision}


The direction of research in IFD points towards finding ways to transfer the success on lab datasets to successful applications on industry datasets \citep{roadmap, potentials_challenges_fututre_directions}.
Both transfer learning and weak supervision can create the opportunity to implement successful algorithms on new data-sets without requiring an expensive labelling process.
Inspired by recent innovations in TLP and NLS, TLS present a third, yet unused direction to integrate the annotations present in CM datasets as labels, learning directly from technical language.
%

%


The potential effects of TLS can be summarised as

\begin{itemize}
  \item Opportunities
    \begin{itemize}
    \item Facilitates direct optimisation on heterogeneous industry data
    \item Methods are available and developed in other research areas
    \item Language data is commonly associated with condition monitoring data-bases
  \end{itemize}
  \item Challenges
  \begin{itemize}
    \item Language annotations are uncertain, and require technical language processing and weak supervision techniques to use
    \item Processing of technical language jointly with industry signals is a novel area of research yet to be developed
    \item Rapid progress requires open industry datasets containing potentially sensitive information
   \end{itemize}
\end{itemize}

%
In this section, we briefly describe the state of TLP and NLS, then combine these into an outline of how TLP can be implemented.

\subsection{Natural Language Supervision}
\label{sec:nls}

Natural Language Supervision \citep{clp_review} is a recent term introduced to describe machine learning optimisation based on free-form text descriptions rather than predefined labels, though language has been used in a similar fashion to labels before.
%
%
\cite{labutov2019learning} trained semantic parsers that interpret questions and feedback from user natural language responses. %
%
%
\cite{hancock2018training}, used natural language explanations of human labelling decision to create BabbleLabble, which converts explanations to noisy labels through a semantic parser.
\cite{murty2020expbert} introduced ExpBERT, which is a BERT variation that forms representations using BERT with natural language explanations of the inputs.

Text-encoding is a crucial part of NLP and has seen rapid development recent years.
Language models \citep{peters2018deep} based on the transformer have increased the representational powers of text encoders drastically \citep{Radford2018GPT1, devlin2019bert, Radford2019GPT2, yang2020xlnet, turingNLG2020, brown2020GPT3}. 
Early examples of text-image pairings used simpler encoding methods, such as Bag of Words(BoW) and TF-IDF, or recursive encodings derived from the word2vec model \citep{mikolov2013distributed}, the predecessor of current transformer-based language models. 
The choice of text-encoder depends on data size and computational power; a larger model can produce better representations, but requires more data and computational power to train. 
Pre-trained language models with general natural language representational capacity, such as BERT \citep{devlin2019bert}, have successfully been fine-tuned on specific tasks with significantly smaller datasets, based on the assumption that the target language and source language has similar underlying distributions.

Optimizing mappings between natural language and images has been done before natural language supervision was introduced; for example, image captioning \citep{image_captioning_survey, image_caption_generation_review, transformer_image_captioning} and visual question answering \citep{Antol_2015_VQA} have both trained mappings between images and text through top-down or bottom-up mappings \citep{anderson2018bottomup} and semantic attention \citep{grained_imgcap, stimulus_imgcap}. 
Knowledge and concepts can also be integrated using language as a supervision tool through neuro-symbolic concept learning \citep{NS_CL}, where visual concepts, word representations, and semantic parsing of sentences are jointly learned.

Image recognition generally uses image-text pairs available from online data crawling to train mappings between text and images.
Learning directly from the text can also facilitate zero-shot classifiers from language descriptions.
\cite{6751432} used text-based descriptions to create a zero-shot image classifier, with text features extracted through Term frequency–Inverse document frequency (Tf-Idf) followed by Clustered Latent Semantic Indexing. 
\cite{lu2019vilbert} introduced ViLBERT, a Vision-and-Language version of BERT, that learns image recognition and language understanding in a two-stream model with interactions between image and text to improve performance compared to single-stream models.
\cite{zhang2020contrastive} classified medical images by utilizing text-image pairs through contrastive visual representation learning (ConVIRT) to learn pairings between images and texts. 
\cite{desai2020virtex} introduced Virtex, which uses captions to enhance pre-training of an image recognition CNN.
\cite{sariyildiz2020learning} mask words in image-annotation pairs to create image-conditioned masked language modelling (ICMLM) for image classification.

In a recent publication, \cite{radford2021learning} at OpenAI presented CLIP, Contrastive Language–Image Pre-training, which popularised the term natural language supervision and showed its efficacy for zero-shot classification. 
They used transformers \citep{transformer} for both text and image encodings \citep{dosovitskiy2020image}, and a contrastive \citep{tian2020contrastive} BoW prediction objective to connect text label to image features in a vector quantised encoding space \citep{oord2018neural, razavi2019generating}.
FILIP by \cite{yao2021filip} uses a fine-grained word-patch image alignment to detect and classify objects based on text descriptions, obtaining finer level-alignment in image-text comprehension through unsupervised natural language supervision.
\cite{jia2021scaling} scaled natural language supervision further by training directly on un-filtered images and annotaions with over one billion image-text pairs.
\cite{wang2021zerolabel} introduced unsupervised data generation to synthesise labels for downstream tasks and thus achieve SOTA results on SuperGLUE \citep{wang2020superglue}.

In earlier models, \cite{6751222} used natural language supervision to train a video event understanding model in 2013 through a rule-based BoW-like model, and \cite{8460937} used language as reward functions for training robots.

\begin{table*}[t!]
    \centering
    \caption{Annotations associated with data from Figure \ref{fig:fault_history}.}
    \begin{tabular}{||l|| c | c ||} 
        \hline
        Case & Months after fault detection & Annotation (translated from Swedish) \\
         \hline 
         \hline
        BPFO indication & 4 & BPFO Env low \\ 
         \hline
        BPFO & 10 &\makecell{BPFO visible in mm/s as overtones \\ high up in the spectrum between \\ 1000 and 2000 Hz. WO written on BPFO } \\ 
         \hline
        Feedback & 12 & \makecell{Bearing replaced YYYYMMDD \\ levels of BPFO low again} \\
         \hline
        \hline
    \end{tabular}
    
    \label{tab:annotations}
\end{table*}
\begin{figure}[h!]
    \centering
    \includegraphics[width=0.5\textwidth]{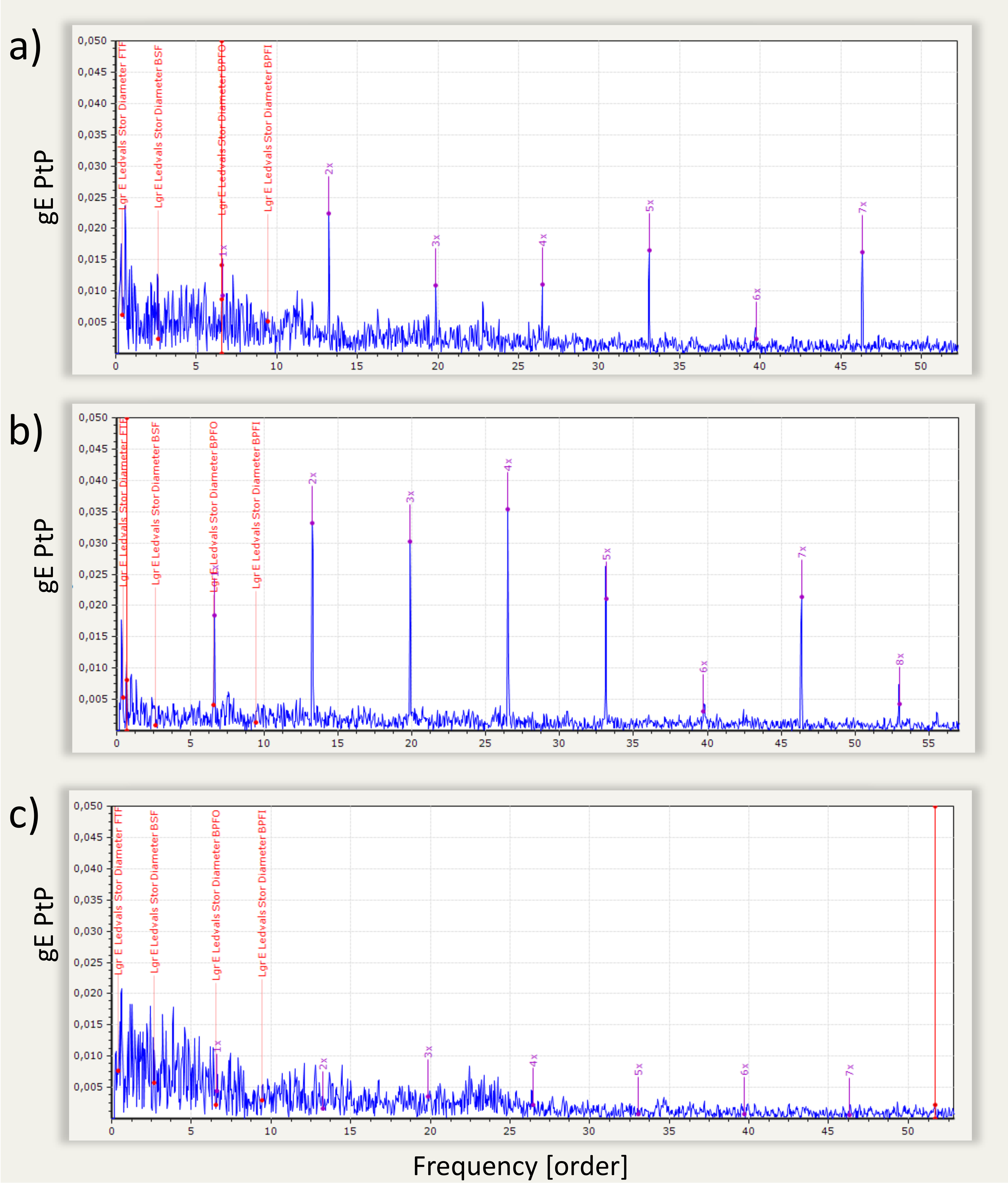}
    \caption{Order analysis results for a vibration signal at a) 4 months; b) 10 months; and c) 12 months after the first indication of a fault in a drying cylinder bearing of a paper machine. Included are also the corresponding text annotations written by experienced condition monitoring analysts employed at the factory. The annotations have been translated from Swedish to English to improve clarity. BPFO peaks are clearly visible in panel a) four months after the first indication of the bearing fault. After ten months, the amplitude of the BPFO peaks in panel b) have increased and a work order (WO) has been written by the analysts. Two months later the bearing has been replaced and no BPFO signature can be seen in panel c).
    }
    \label{fig:fault_history}
\end{figure}

\subsection{Technical Language Processing}
\label{sec:tlp}
The term Technical language processing was introduced in Dec 2020 by \cite{technical_language_processing} in collaboration with the American National Institute of Standards and Technology, and concerns the application of NLP techniques and pipelines on technical language.
The processing of technical language requires natural language processing methods with additional considerations related to the characteristics of technical language, which is characterised by a higher frequency of information-rich key-words, more abbreviations, and considerably less data than natural language.
TLP can be used as a basis for TLS, but can also directly enhance CM practices by offering insights into key performance indicators from work order features \citep{TLP_KPI_MWO_2021}.

The challenges inherent in using free form text data from industrial contexts - namely data scarcity, a high density of important but (to the model) undefined abbreviations, and technical terms and concepts critical for maintenance context but not inherently defined by their context - are different enough from current NLP research to warrant its own key word in TLP.
An ideal TLP model which performs as well as modern language models do on natural language would be able to answer free-form questions on the text dataset, understanding what parts of MWOs and annotations indicate fault class, severity or maintenance actions, and similar tasks currently only possible with human analysis.
However, the aforementioned challenges make direct implementation of pre-trained language models difficult; \cite{dima2021adapting} describe the challenges in adapting natural language processing for technical text in detail. and warn against possible shortcomings of implementing SOTA NLP models without considering the specific needs of the process or the people involved.
A large black box model can lead to issues with model justifiability, scrutiny and bias, undermining confidence in the system.

\subsubsection{Technical Language Processing Implementations}
Implementations of word embedding models, among those language models, have seen some testing.

\cite{nandyala_TLP_embeddings}, implemented five models for vector representation of technical text using an open source dataset
describing 5,485 work orders for 5 excavators \cite{Hodkiewicz2017WhyAA}.
To evaluate their results they relied on qualitative human evaluation in word and sentence similarities, as well as word cluster projections, as no obvious extrinsic evaluation tasks are available in the model.
The authors also survey the literature on fields with challenges similar to those faces in technical language, and discovered similar problem formulations in finance, law, medicine and bio-medicine.
In particular, the bio-medical community has developed public datasets for training and benchmarking of domain-specific NLP models.

\cite{camembert} used a French version of RoBERTa \citep{liu2019roberta} called CamemBERT to estimate language features such as duration and criticality of maintenance problems based on operator descriptions.
They used equipment descriptions, importance and symptoms as input, and type of disturbance as criticality output (dominant or recessive) and maintenance workload (hours) as outputs for duration.
Such input-output pairs allow for extrinsic evaluation, but also fine-tuning of model parameters.
The results indicate that Tf-IDf considerably outperforms the base CamemBERT and almost as well as fine-tuned CamemBERT, which implies that the task, data or evaluation are insufficient to fully benefit from the representational capacities of large language models.

\cite{brundage_TLS} show an association between signal values and expert annotations by generating a technical language dataset with the help of two technicians.
One technician generated and monitored faults, followed by another technician writing annotations.
The authors find a clear correlation between annotation contents and expert condition monitoring, which presents a strong case for language supervision.
\cite{myself_substitution} investigate the effect of out-of-vocabulary technical terms on BERT and SentenceBERT performance annotation representations by substituting key terms with in-vocabulary natural language terms.
The challenges of evaluation without labels or benchmark datasets were also discussed, and two methods to simulate extrinsic metrics were suggested.
The authors found that the clusterability as measured by k-means score, and the predictability of automatically assigned fault class labels, both improved with only a few key words substituted.

\subsubsection{Technical Language Processing Embeddings}

Language-based models require a mathematical representation of language.
This is achieved through pre-processing and an embedding algorithm.
The pre-processing step involves tokenisation, cleaning and spell-checking, stop-words removal, stemming/lemmatisation, and fundamental language analysis such as part of speech tagging and named entity recognition.
The embedding algorithm can be as simple as one-hot encoding or a complex massive transformers based architecture.

Figure \ref{fig:fault_history} and Table \ref{tab:annotations} illustrate an example of technical language annotations and condition monitoring signals from a craft liner production plant in northern Sweden.
The Figure shows three different envelope-filtered measurements associated with the annotations shown in the Table.
The first annotation indicates that there is a fault of class Ball-Pass Frequency Outer ring (BPFO) with a low severity, which is related to the low-intensity peaks at characteristic kinematically based order frequencies in the spectrum.
The second annotation describes that the corresponding overtones have increased in magnitude and that a work order has been written.
At that point the fault is estimated to be more severe and at the end of its RUL, so the component (bearing) has to be replaced.
Finally, the third annotation informs that a bearing has been replaced and that the vibration levels are low, indicating a healthy component.

\subsubsection{Challenges and Solutions}
Pre-processing of technical language faces several difficulties, as use of technical language can vary even in the same field, and there is no uniformly defined list of stems/lemmas, stop-words or correct spellings.
For instance, if a CM dataset contains faults of class "Ball-Pass Frequency Outer" (BPFO) and "Ball-Pass Frequency Inner" (BPFI), but one is considerably more common than the other, an automated spell-checker might assume that one is a spelling error.
Likewise, there is no defined dictionary for stemming of technical words such as BPFO or BPFI, and reducing both words to "BPF" naturally loses critical information.
Therefore it is necessary with a "human-in-the-loop" system until a level of language processing maturity which accurately covers the heterogeneous field of technical language is achieved.
One dictionary of technical stop words has been produced \citep{stopwords_TLP_2021}, though it is not necessarily the case that this list is accurate for industries besides those covered in the article.

Encoding technical language to vectors faces a major challenge in that many technical words specific to industries are not in the vocabulary of NLP models trained on natural language.
Addressing this directly with NLP methods is thus related to handling out of vocabulary (OOV) words.
A common method to deal with OOV words, used in for instance BERT \citep{devlin2019bert} and GPT \citep{gpt1, gpt2, gpt3}, is to input subword encodings such as byte-pair encodings (BPE) \citep{Gage1994BPE, sennrich2015neural} or WordPieces \citep{wordpiece_2012, Wu2016GooglesNM}, rather than the words themselves as inputs to the model. 
Both models work by learning to maximise the coverage of words in the corpus using a typically fixed amount of subwords.
Thus, common words are assigned one whole token, while uncommon words or word endings, such as the "ing"-suffix in for instance "running", might be assigned multiple tokens.
The difference between BPE and WordPiece comes mainly from how the subwords are assigned, where BPE chooses the most frequent byte pair and WordPiece chooses the the pair which maximises the likelihood of the training data.
Other models try to learn to predict the meaning of an unknown word based on surrounding words, individual characters, or a combination of both \citep{OOV_DL_2020}.
Implementing an OOV solution which allows transfer learning of a pre-trained deep learning NLP encoder could potentiate more semantically accurate representations of technical language word embeddings, which in turn would improve the potential for TLS.

Another method to encode technical language is through human designed expert systems - essentially a set of rules describing the keywords for faults, actions, severities etc \citep{TLP_keyword_extraction_2018}.
The annotation $$"High \; BPFO \; in \; env3. \; WO \; on \; bearing \; replacement"$$ would thus be decomposed into $$class - BPFO; \;severity - high; \;detected \;in - env3; $$ $$\;action - WO \;replacement; \;action \;target - bearing.$$
These keywords can then serve as targets for annotation prediction or language based supervision, acting as less noisy labels than learned embeddings for language representations.
However, such a system is difficult to scale and vulnerable to new keywords being introduced, essentially requiring tailored engineering and maintenance for each unique industry.
It is also vulnerable to oversights from the engineers of the expert system, for instance missing negations in statements, unforeseen keyword usage or a lack of context due to the removal of semantics.


\subsection{Outline of Technical Language Supervision concepts and model}

In the infant stage of TLP, classical NLP methods such as stop-word removal, lemmatisation, stemming and BoW analysis have been used.
A potential improvement is to apply more recent innovations in pre-processing and analysis, such as word embedding algorithms coupled with manual tagging of industry-specific technical language.

\begin{figure}[t]
    \centering
    \includegraphics[width=0.5\textwidth]{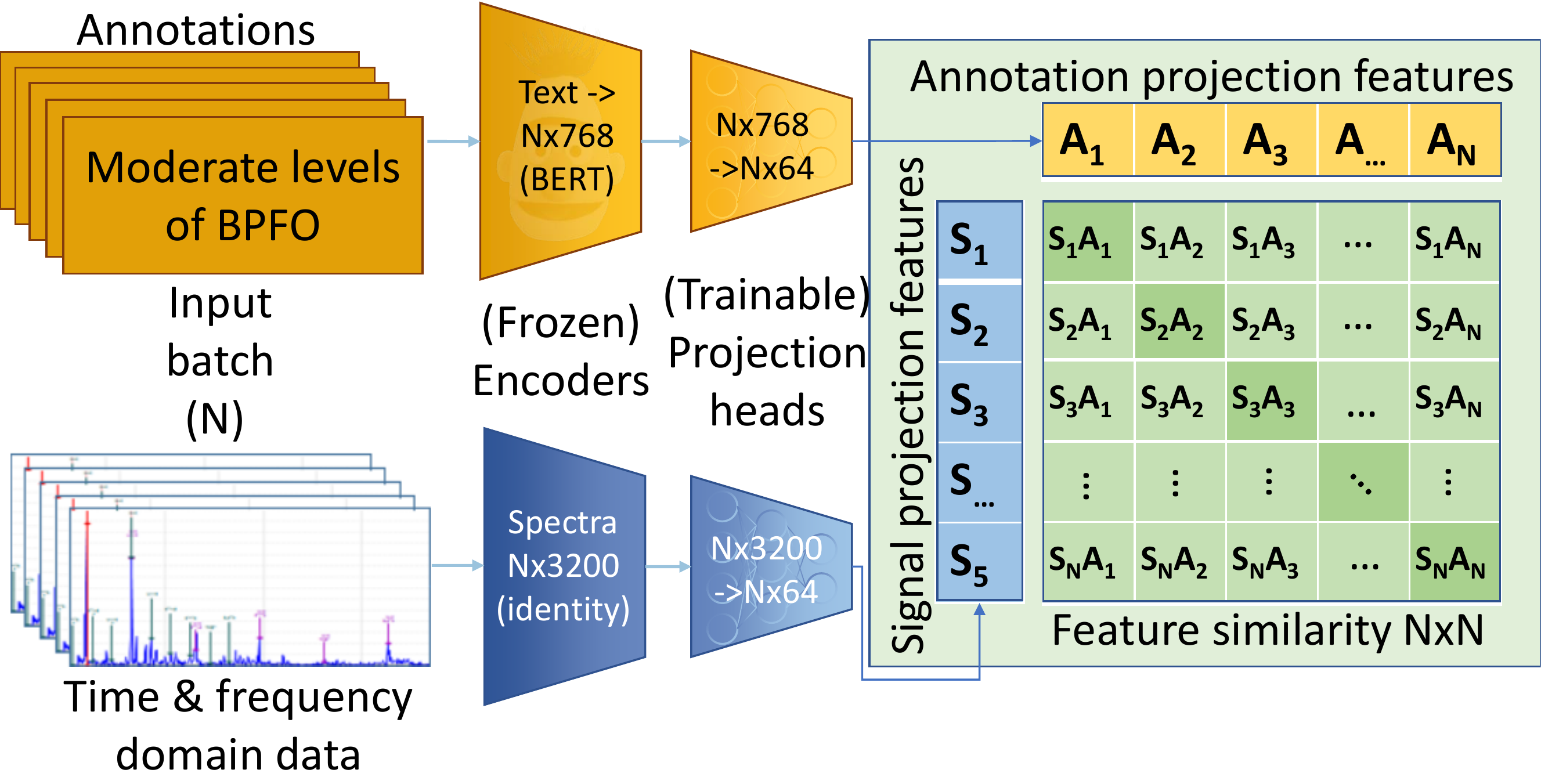}
    \caption{Example illustrating the pre-training step of a natural language supervision model. Annotations and time-frequency domain signal features are encoded, and the model is optimised to connect the correct text-feature pair in the batch of training examples, here marked with dark green colour, through contrastive learning.}
    \label{fig:contrastive}
    \vspace{-2mm}
\end{figure}

\begin{figure}[h!]
    \centering
    \includegraphics[width=0.51\textwidth]{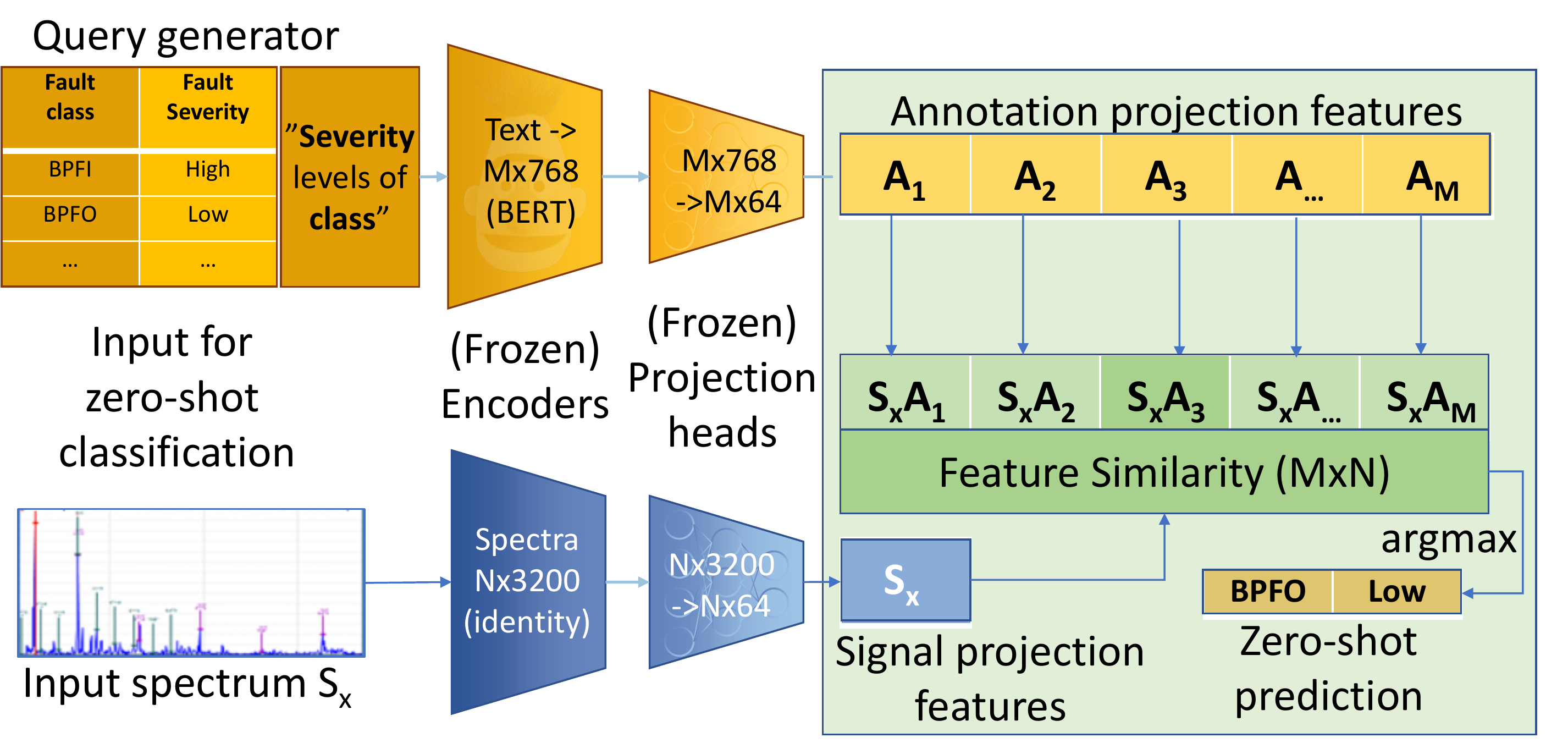}
    \caption{Example illustrating how inference can be generated with the natural-language supervised model outlined in Figure \ref{fig:contrastive}. Signal and language inputs are compared in the projection space learned during pre-training, and pairs with the highest feature similarity are used as outputs in either spectrum retrieval or zero-shot classification.}
    \label{fig:zero-shot}
    \vspace{-2mm}
\end{figure}

%


Figures \ref{fig:contrastive} and \ref{fig:zero-shot} show a TLS model inspired by the CLIP model \cite{radford2021learning} describing natural language supervision.
In the pre-training step, a technical language encoder and a fault diagnosis encoder are used to produce fault and text features.
A mapping between fault and text encodings is learned through contrastive learning \citep{tian2020contrastive, zhang2020contrastive}.
In the inference phase, the same encoders are used, but additionally there exists a label query mechanism that maps an input signal to the annotation-based label that is closest to the query in the joint data and language embedding space.

In the case of IFD of rotating machinery, the input is typically sensor data in time-, frequency- and time-frequency-domains.
IFD data encoding methods are described in section II, and typically consist of variations of CNNs. 
Recently, the Transformer \citep{transformer}, an architecture introduced to model long-range dependencies and training inefficiencies in NLP, has been successfully used for image recognition without any convolutions in the model \citep{dosovitskiy2020image, wu2020visual_transformer, han2021survey_visual_transformer}.


In order to train classification or regression models using language, and not just an annotation generator, a language based labelling method is required. 
Based on current state-of-the-art methods, some human intervention is required in this step to pre-define the label-space, so that annotations can be mat-ched to the closest label semantically.
In \citep{radford2021learning}, a BoW method is implemented to complete pre-defined sentence structures by inserting the correct term chosen from the bag.
A similar model could be used in IFD, with more than one degree of freedom in the query to label both fault class and severity
Potentially, further degrees of freedom also enables labelling time-aspects of fault evolution. 
With a large text dataset and access to well defined labels in parallel with the annotations, a mapping between a more feature-rich encoding and the label space can be learned and implemented to produce labels in a weakly supervised manner for data-annotation pairs where labelled data are not available.

In the case of CM data, the volume and density of text data is low compared to web-crawl results for captioned images on the Internet, or extensively annotated datasets such as COCO \citep{lin2015microsoft_coco}.
The language is also domain specific, and annotations are connected to time-frequency data recordings in the dataset, while the semantics of an annotation can be based on analysis of trends over many measurement recordings.
This motivates the use of pre-trained models, in combination with feature-engineering and fine-tuning to adapt the model to the domain-specific terms used in process industry. 
Weak supervision will also be required to deal with unannotated faults, time-delays, a lack of annotations in healthy data, and noise in the annotations resulting from domain-specific language, spelling errors, and grounding noise due to subjective interpretations.

\section{Case Study}
We implement a version of the architecture presented in Figures \ref{fig:contrastive} and \ref{fig:zero-shot} using data from a craft paper production plant in northern Sweden, with spectrum and annotation embedding projection heads as trainable parameters through contrastive learning.

\subsection{Data}

\begin{figure}[b]
    \centering
    \includegraphics[width=0.49\textwidth]{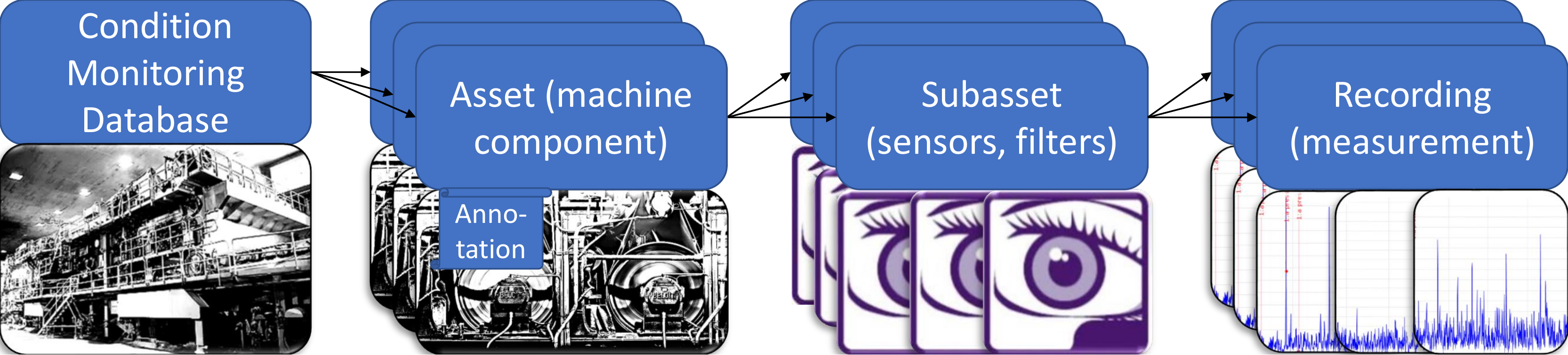}
    \caption{Schema of data structure in a condition monitoring database. The database consists of multiple machine parts called assets, which sometimes have associated annotations. Each asset has multiple subassets consisting of different types of signals, from one or more sensors. Each subasset consists of multiple recordings in the form of time series and spectra measurements.}
    \label{fig:schema}
\end{figure}

The data used comes from six months of recorded data in two large paper mills producing Kraftliner in northern Sweden, and consists of annotated condition monitoring signals from assets, such as dryers, rollers and gearboxes etc.
%
%
Figure \ref{fig:schema} shows a schema of the data structure for each paper mill.
Each paper mill forms a database.
The database consists of multiple machine parts called assets, which occasionally have associated annotations when faults have been detected and diagnosed.
Each asset has multiple subassets consisting of different types of signals, from one or more sensors.
Subassets can be two sensors mounted on the same asset but at opposite ends, or signals from the same sensor that have been transformed using filters such as enveloping.
Subassets consist of multiple recordings in the form of time series and spectra measurements, which are the data used in this case study.
A recording is one series of data, typically 6400 vibration measurement samples taken over 6.4 seconds, from which spectra with 3200 samples up to 500Hz are computed.
Thus, for each annotation there is one associated asset, with multiple subassets, with multiple recordings.

%
In our dataset, we have 109 annotations with a total of 21090 associated recordings present in a span of ten days before and after the annotation date.
Many annotations are identical, and of the 109 annotations there are 43 unique fault descriptions.
As data scales up, the number of unique annotations will also increase, which is why a pre-trained language model is needed to ensure system scalability.

\subsection{Text Encoder}
The text encoder part of out case study TLS model is seen in Figure \ref{fig:contrastive}, shown in orange at the top of the figure.

The annotations are embedded using a pretrained and frozen SentenceBERT \citep{reimers2019sentence} model trained on Swedish corpora \citep{rekathati2021swe_sentence_bert}, which transforms every annotation to a 768-dimensional embedding vector. as shown in the first two boxes.
SentenceBERT is based on BERT and RoBERTa, but is trained to specifically produce good sentence embeddings through siamese and triplet networks \citep{facenet_2015}.
In the normal BERT model, each word is projected to a 768-dimensional embedding vector.
For example, an annotation with ten words is embedded with dimensions 10x768.
%
%
To use these embeddings for downstream tasks, it is common to pool them to 1x768 then use a feed-forward neural network (FFN).
Pooling can be accomplished by averaging each embedding, taking weighted max values, or by using the classification (CLS)-token, which is a final token added to the BERT model that effectively becomes a learned pooling of the self-attention.
SentenceBERT is a BERT-based model fine-tuned on the task of pooling word embeddings to sentence-embeddings, using corpora with similar and dissimlar sentences, and an objective function defined to minimise some distance measure, either softmax, cosine or euclidean between triplets, between similar sentence embeddings.
Thus, annotations, which typically consist of one sentence, can be transformed directly to 1x768 with a model specifically optimised for this task.
%

An FFN is then used to reduce the dimensions down to 64 to introduce trainable parameters and reduce the complexity of the dot-product in the contrastive learning step.
The FFN is a simple two-layer network with one skip-connection going from 768 to 64 to 64, with a Gaussian error linear unit (GELU) activation function, a 10\% dropout, and layer norm.
The output of the FFN is then used as input for the contrastive learning step, seen in the rightmost two boxes of the figure

\subsection{Signal Encoder}
As there are only 109 annotations it is challenging to optimise a network at the asset or subasset level. 
Therefore we propagate the labels down to the recordings level, where we have 21090 spectrum-annotation-pairs with 43 unique annotations.
Thus, the same annotation at an asset will describe every spectra related to that asset, even if some spectra are void of fault features.
%
%
However, as shown by \cite{jia2021scaling} and \cite{simvlm}, noisy text-image pairs can still converge to a general understanding through the weak supervision that is still present, and it is likely that the same will hold true when replacing images with sensor data.

We directly use the spectra as the fault features, which can be interpreted as the pre-trained model being a FFT and envelope filter of the raw time series.
%
%
The spectra are projected from 3200 to 64 dimensions with the same reasoning and the same model setup as the annotation embeddings.
This is shown in Figure \ref{fig:contrastive} as the spectra encoder being empty, going from 3200 to 3200.
As with the annotation embeddings, the resulting 64-dimensional vectors are then sent to the contrastive learning step, seen in the next blue box.

\subsection{Contrastive Learning}

\begin{figure}[t]
\vspace{-1mm}
    \centering
    \includegraphics[width=0.5\textwidth]{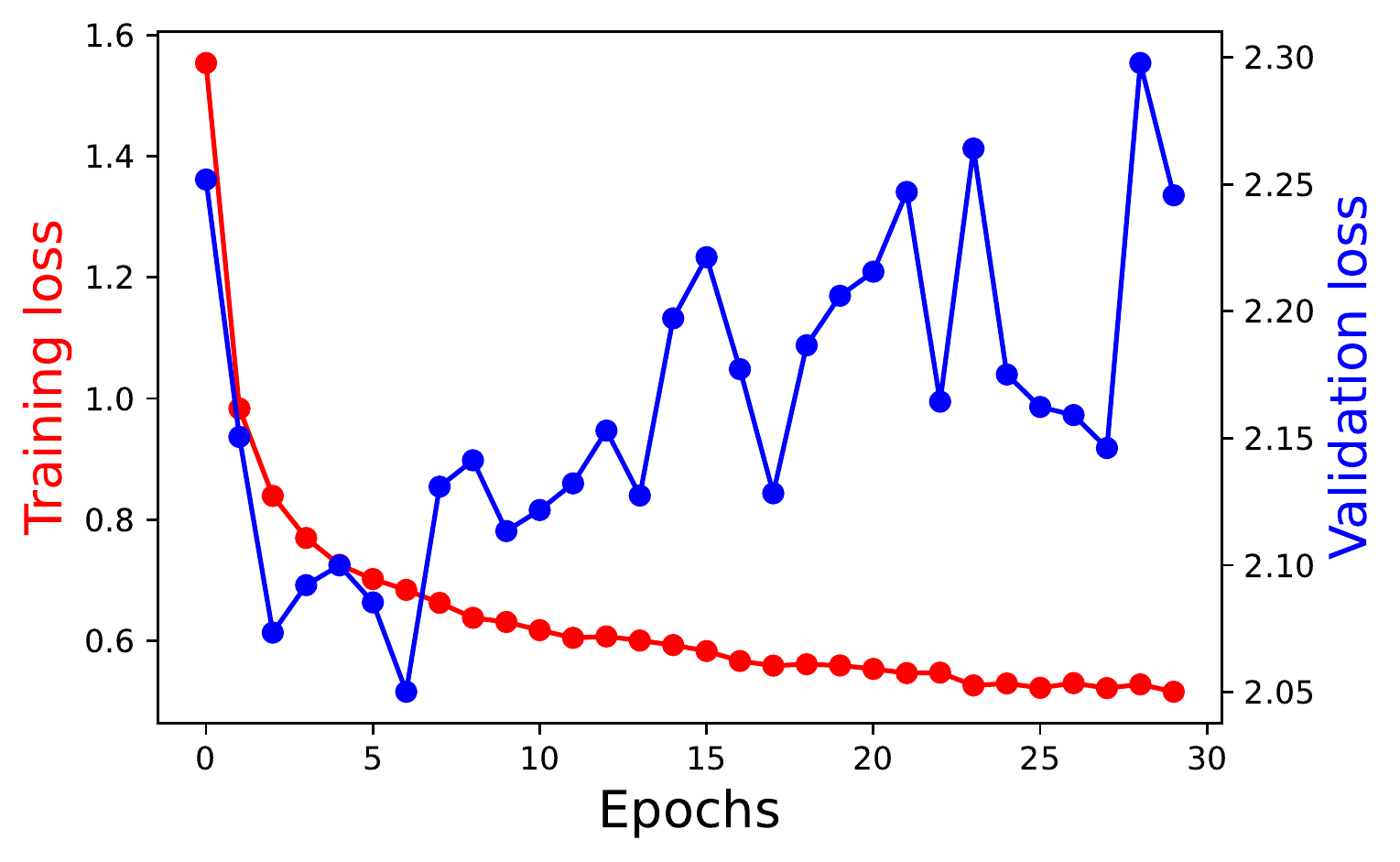}
    \vspace{-5mm}
    \caption{Training and validation contrastive loss for case study model.}
    \label{fig:epochs_loss}
    \vspace{-2mm}
\end{figure}

We train the data using contrastive loss to project positive pairs of signals close and negative pairs further away in a projection space, inspired by the methodology presented in \citep{radford2021clip}.
Logits are computed through the dot product of the text and spectrum embeddings in a batch.
The self-similarities of spectrum and text embeddings are then computed through the dot product with themselves.
The targets, the "labels" for the constrastive loss, are then computed as the softmax of the averages of the self-similarities.
The loss for the text-encoder and the spectrum-encoder are then computed separately through cross entropy loss of the logits and the targets.
Finally, the model batch loss is defined as the mean of the spectrum and text losses.
We trained the model for only three epochs, as the loss on the validation set quickly started deviating from the train loss, as seen in Figure \ref{fig:epochs_loss}.
%

\subsection{Zero-shot analysis}
%
Finally, the pre-trained model is used to show which spectra in the dataset best correspond to fault queries through spectrum retrieval, and to predict fault classes based on an unlabelled spectra with a label query through zero-shot classification.
More specifically, unlabelled spectra chosen from the dataset, and manually chosen label queries, are both used as inputs, while the highest dot product of the embeddings generates a prediction output.

\begin{figure}[t]
    \centering
    \includegraphics[width=0.45\textwidth]{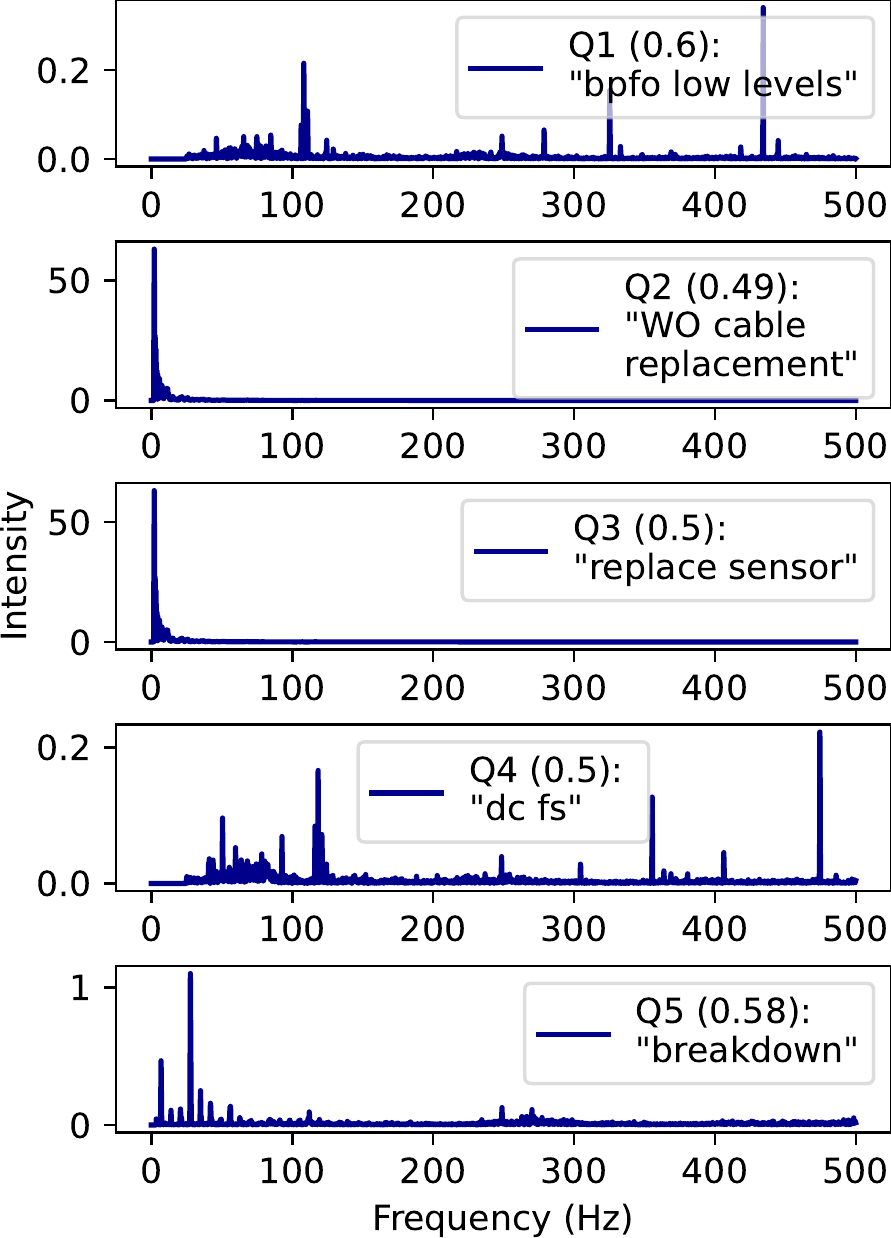}
    \caption{Spectrum retrieval using text queries to sample the top three spectra with the highest embedding dot products}
    \label{fig:image_retrieval}
\end{figure}

\begin{table}[b]
    \centering
    \vspace{-2mm}
    \caption{Query inputs for spectrum retrieval and zero-shot classification}
    \vspace{-2mm}
    \begin{tabular}{|c|c|}
    \hline
       Query ID & Query  \\
       \hline
       Q1  &  "BPFO low levels"  \\
       Q2  &  "WO cable replacement" \\
       Q3  &  "Replace sensor" \\
       Q4  &  "DC FS" \\
       Q5  &  "Breakdown" \\
       \hline
    \end{tabular}
    \label{tab:queries}
\end{table}

\begin{figure}[t]
    \centering
    \includegraphics[width=0.5\textwidth]{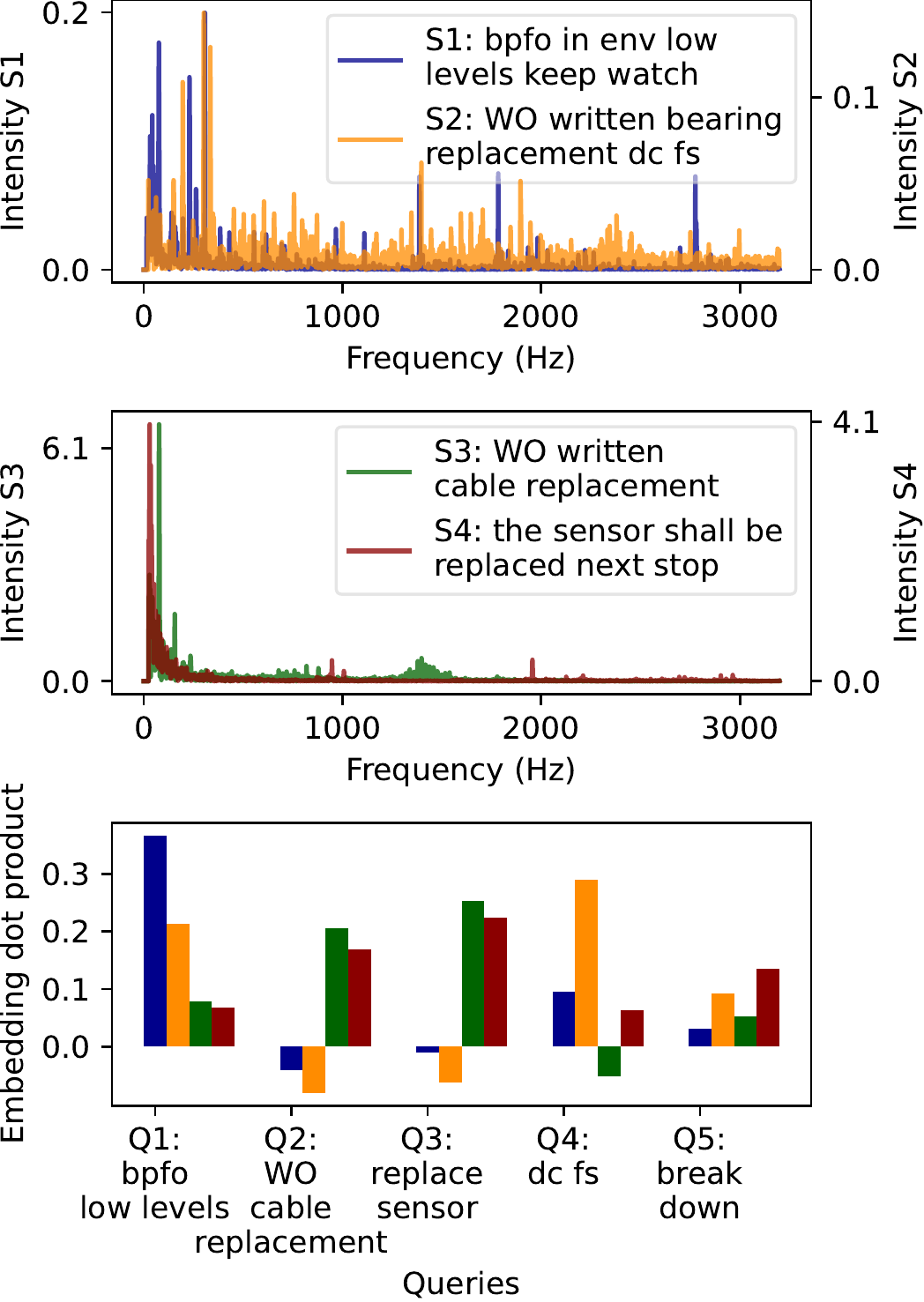}
    \caption{Zero-shot implementation on the case study data. Input spectra are shown in the upper two graphs, and input queries with matching inner products are shown in the lower graph.}
    \label{fig:query_zero_shot}
\end{figure}

Figure \ref{fig:image_retrieval} shows spectrum retrieval using queries, described in Table \ref{tab:queries}, as inputs and receiving matching spectra as outputs.
The queries are embedded using the pre-trained technical language supervision model, alongside all spectra in the training and validation set.
The output spectra are those with the highest embedding dot product.
%

Figure \ref{fig:query_zero_shot} illustrates a zero-shot classification implementation of the technical language supervision framework, with queries also described in Table \ref{tab:queries}.
Four examples of spectrum inputs are shown in overlapping pairs in the upper two parts of the figure.
The corresponding annotations and axes for these spectra are colour coded and marked as S1-S4.
The lower part of the figure illustrates zero-shot classification with five queries, where the inner product between the queries and the spectra was computed.
The inner product between a spectra and a query is represented directly over the query, with colours indicating which spectra the bar is related to.
%

%
%
%
%
%
%
%
%

\subsection{Results}
\subsubsection{Spectrum retrieval}
The results of the case study indicate that even with a limited amount of data and a relatively simple model with few hyperparameters, there are aspects of fault diagnosis learned without any labels.
For instance, the top four spectra chosen in the spectrum retrieval task shown in Figure \ref{fig:image_retrieval} are all examples of signals that correspond to their respective query;
%
%
%
queries 1 and 4 retrieve spectra indicating bearing faults, with high frequency peaks likely corresponding to characteristic frequencies of bearings, while queries 2 and 3 both indicate cable or sensor faults, seen in the unnaturally high intensity close to zero, indicating a bias in the time series.
Query 4 illustrates one interesting property of the correlations between language and signals, where "DC FS" means "drying group free side", which is a phrase commonly seen in conjunction with bearing fault detection or bearing replacement work orders.
Query 5 was chosen to test the model where no clear correlations were to be expected, as there were very few occurrences of breakdown in the dataset and breakdown does not have one clear signal representation.
However, upon consultation with an expert analyst, we learned that the model had picked up a correlation between spectra indicating looseness, which apparently was the reason behind this breakdown.
Thus, the analyst found this spectrum retrieval valuable and successful, indicating the need for close collaboration with industries even when developing self-supervised data-driven models.

\subsubsection{Zero-shot predictions}
The zero-shot predictions shown in Figure \ref{fig:query_zero_shot} produce good results, where Q1 and Q4 correctly have a higher values for spectra 1 and 2 than for spectra 3 and 4, while Q2 and Q3 correctly correlate more with spectra 3 and 4.
In particular, Q1, "BPFO low levels", correctly correlates significantly more with the spectra whose annotation reads "BPFO in env low levels keep watch", and Q4 likewise correctly correlates much more with S2.
Furthermore, both queries correlate more with S1 and S2 which are spectra that indicate bearing faults, despite the individual characteristics of each spectra being different.
Q2 and Q3 both correlate strongly to the similar spectra S3 and S4, with Q3 correlating stronger with both spectra, and S3 stronger with each query.
However, since cable and sensor faults in general show similar feature spaces, both queries are accurately mapped to similar spectra.
Q5, "breakdown", correlates poorly with all chosen spectra, which is an accurate classification as none of the input spectra should indicate a breakdown.

In both the spectrum retrieval and the zero-shot prediction we used normalised text embedding and unnormalised spectrum embeddings before normalising the dot product, as opposed to the normalised spectrum embeddings used during training.
Normalising the text embeddings had little impact on either task, but the spectrum retrieval was affected considerably by normalisiation of the spectrum embeddings, producing better retrievals with higher values for BPFO-related annotations, but lower values and worse retrievals for cable and sensor faults, while zero-shot predictions were relatively unaffected.

\subsection{Discussion}
The technical language supervision model outlined and implemented in this case study is a basic adaptation of the model used in \cite{radford2021clip}.
It faces several challenges related to the application to technical language and condition monitoring signals, which are discussed in the following subsections.
Table \ref{tab:discussion_table} summarises the tasks, challenges and proposed approaches for text encodings, signal encodings, contrastive learning and zero-shot classification.

\begin{table*}[t]
    \centering
    \caption{Different TLS tasks, challenges and approaches}
    \begin{tabular}{||l|| c | c ||} 

            \hline
        Task & Challenges & Approaches \\
            \hline 
            \hline
        \makecell{Encoding \\ technical \\ language.} & \makecell{Technical language different \\ from natural language. \\ Limited data availability.}  & \makecell{Technical language processing, see \ref{sec:tlp}. \\ Self-supervised pre-training, see \ref{sec:nls}. \\ Supervised fine-tuning, see \ref{sec:nls}.}\\
            \hline
        \makecell{Encoding \\ fault \\ features} &\makecell{Labelled industry data scarce. \\ Lab features difficult to transfer. \\ Non-linear evolution of fault severity. \\ Fault severity levels industry specific.} & \makecell{Transfer Learning, see \ref{sec:transfer_learning}. \\ Weak Supervision, see \ref{sec:weak_supervision}. \\ Contrastive learning for fine-tuning, \\see \ref{sec:inexact} and \ref{sec:nls}.} \\ 
            \hline
        \makecell{Contrastive learning \\ optimisation.} & \makecell{Faults appear and evolve over \\ multiple recordings and signal types.}  & \makecell{Sequential model projection \\ heads, see \ref{sec:contrastive_learning}. \\ Data augmentation, see \ref{sec:contrastive_learning}. }\\
            \hline
        \makecell{Evaluating zero-shot. \\ performance and implementation.} & \makecell{Novel task. \\ No benchmark test set.} & \makecell{Industry expert analysis \& \\ Industry test deployment, \\ see \ref{sec:zero_shot_predictions}.} \\
         \hline

    \end{tabular}
    
    \label{tab:discussion_table}
\end{table*}

\subsubsection{Text encoder}
The main challenge for the text encoder is to create good embeddings of technical language, as they are the basis for the potential of the contrastive learning step.
As discussed in Section 3.2, this challenge is due to technical language being different from the natural language normally used to train language models, and technical language data scarcity.
In this case study, we opted to use a pre-trained natural language model without any fine-tuning.
Three approaches for improvements of technical language encodings are shown in the table, which can be summarised as using small-data industry specific solutions through technical language processing, discussed in Section \ref{sec:tlp}; large data self-supervised pre-training solutions; and supervised fine-tuning, both discussed in Section \ref{sec:nls}.

%

\subsubsection{Signal encoder}
For the signal encoder, the main task is to produce good fault feature representations prior to the projection head, comparable to the language model step of the language encoder.
%
%
%
%
The lack of labelled industry data sets, the difficulty of feature transfer and the non-linear and industry-specific properties of fault severity, are all challenges for this task.
Approaches to overcome these challenges are discussed in Section two, but more specifically transfer learning, weak supervision and contrastive learning are viable approaches, with specific sections shown in Table \ref{tab:discussion_table}

%
%
%
%
%

\subsubsection{Contrastive learning}
\label{sec:contrastive_learning}
The task of the contrastive learning part of the model is to force positive pairs to a similar projection space, while negative pairs are pushed away.
The main challenge in this step is related to data properties, where annotations are too scarce to fully leverage the utility of scale that is shown in NLS \citep{NLS_review}, and fault evolution too nonlinear for annotation propagation to accurately work as data augmentation.
Furthermore, 
%
unlike in NLS prediction of image classes, TLS individual recordings are insufficient information to fully assess fault characteristics, akin to describing a movie from just one frame.
Thus, multiple recordings must be considered to mimic human analysis in the contrastive learning step, which requires methods able to attend to sequential data such as recurrent neural networks or transformers, either as projection heads or integrated in the text and signal encoders.

Propagating annotation embeddings to each corresponding recording increases the size of the dataset, but also leads to inaccurate supervision from annotations on the recordings level, arising due to the inexactness between recordings level and asset level.
For example, if a sensor is faulty at half of its measurements, but works for the other half, the model should ideally be trained only on the faulty signals.
Likewise, BPFO is typically detected first in envelope spectra, thus resulting in BPFO annotations being associated with normal spectra where BPFO features have likely not appeared yet.
The variety of input types in the spectra inputs is in itself an issue for optimisation, as the network will have to learn to project two very different signals to the same projection in the joint embedding space.
However, knowledge of expected fault behaviour with regards to annotation types could be leveraged to perform improved data augmentation and more accurately propagate annotations in time with annotation contents changing depending on fault type and time distance from true annotation.
 
\subsubsection{Zero-shot predictions}
\label{sec:zero_shot_predictions}
The main challenges with zero-shot classification in an industry environment is that it is a novel field and hard to evaluate without labelled test sets.
The contrastive loss or accuracy during optimisation is relative to the model, and offers little insight into model performance at implementation.
Therefore we use Figures \ref{fig:image_retrieval} and \ref{fig:query_zero_shot} to illustrate model performance for two test scenarios.
This evaluation requires prior fault diagnosis knowledge however, compared to the much simpler task of evaluating natural language supervision classification for image captioning.
However, the efficacy of the model can also be evaluated by test deployment in industry, where feedback from industry experts evaluates whether the model works to improve current fault diagnosis practices or not.

Investigating the zero-shot predictions in Figure \ref{fig:query_zero_shot} showed that a spectrum containing BPFO features gave high inner products also for cable and sensor queries, and we speculated that this might be due to latent BPFO features occasionally seen in the cable and sensor-associated spectrum training data.
This is an issue of incomplete supervision, which is further exacerbated if unannotated data is used during training, as the absence of fault annotations does not necessarily guarantee the absence of fault features, given that early faults might go undetected by current analysis.
The issues of weak supervision can be addressed by adding data-specific solutions, by for instance limiting extraction dates to after the annotation, or adding pre-processing of annotations to manually handle "replaced"-like annotations as a different class.
This issue might also be solveable by simply scaling up data, which has worked in natural language supervision as discussed in Section 4.3.

\section{Conclusion}
The fault descriptions and maintenance records commonly stored in modern process industry CM systems are unexploited sources of information for training IFD systems.
Language present in CM datasets can be used for technical language based supervision of IFD models to facilitate automation of routine FD tasks and develop more accurate decision support for complex tasks 
\citep{sais_paper}.
Since language-based labels are intrinsically uncertain, weakly supervised learning methods need to be developed, which can also support transfer learning of pretrained IFD models with labels extracted from language in industry datasets.
Our experiments show that even with a basic TLS implementation, without custom signal processing or pre-trained fault diagnosis encoders, a joint embedding space for annotations and fault features can be learned and used for zero-shot classification.

Improvements in TLS can occur both through an enhancement of the TLP pipeline for technical language representations, or through augmented integration of IFD-based signal encoders.
However, a major challenge for TLP and TLS research is the
%
lack of realistic and open
annotated industry data, which can be 
%
used for comparative studies and benchmarks.
Furthermore, the assistance of industry experts was sometimes required to understand the annotation language and how annotations were motivated by signal features and the context.
%
Thus, in this work the collaboration between industry and academia was key.
Open access annotated datasets with clearly described features and valid benchmark tasks are needed to make this important direction of research more readily accessible.

\section*{Acknowledgement}
This work is supported by the Strategic innovation program Process industrial IT and Automation (PiIA), a joint investment of Vinnova, Formas and the Swedish Energy Agency, reference number 2019-02533.

The analysis of the results was done with help from Håkan Sirkka, a condition monitoring analyst with experience of the industry data used.

We thank the members of the project reference group including Per-Erik Larsson, Kjell Lundberg, Håkan Sirkka and Peter Wikström, for valuable inputs.

KL thanks Prakash Chandra Chhipa for helpful discussions on contrastive learning and joint embedding spaces.


\bibliographystyle{apacite}
\PHMbibliography{references}


\end{document}